\documentclass[10pt,twocolumn,letterpaper]{article}

\usepackage{cvpr}              
\definecolor{cvprblue}{rgb}{0.21,0.49,0.74}
\usepackage[pagebackref,breaklinks,colorlinks,allcolors=cvprblue]{hyperref}

\usepackage[accsupp]{axessibility}  
\usepackage{graphicx}
\usepackage{booktabs}
\usepackage{amsfonts}
\usepackage{amsmath}
\usepackage{pifont}
\usepackage{multirow}
\usepackage{subcaption}
\usepackage{enumitem}
\usepackage{makecell}
\usepackage{comment}

\newcommand*{\schname}{EMDUL}

\def\paperID{} 
\def\confName{CVPR}
\def\confYear{2026}

\title{Expanding mmWave Datasets for Human Pose Estimation with\\
Unlabeled Data and LiDAR Datasets 
}

\author{Zhuoxuan~Peng 
\hspace{1em}
Boan~Zhu 
\hspace{1em}
Xingjian~Zhang 
\hspace{1em}
Wenying~Li 
\hspace{1em}
S.-H.~Gary~Chan\\
The Hong Kong University of Science and Technology\\
{\tt\small zpengac@cse.ust.hk, \{bzhual, xzhangha, wlidt\}@connect.ust.hk, gchan@cse.ust.hk}
}

\begin{document}
\maketitle
\begin{abstract}

Current 
millimeter-wave (mmWave) 
datasets for human pose estimation (HPE) are scarce and lack diversity in both point cloud (PC) attributes and human poses, hindering the generalization ability of their trained models. 
On the other hand, unlabeled mmWave HPE data and 
diverse LiDAR HPE datasets are readily available.
We propose \schname, 
a novel approach to {\bf e}xpand the volume and diversity of an existing {\bf m}mWave {\bf d}ataset using {\bf u}nlabeled mmWave data and
{\bf L}iDAR datasets.
\schname{} consists of two independent modules, namely a pseudo-label estimator to annotate  unlabeled mmWave data, and 
a closed-form converter that translates an annotated LiDAR PC to its mmWave counterpart.
Expanding the original dataset with both LiDAR-converted and pseudo-labeled mmWave PCs,
schname{} significantly boosts the performance and generalization ability of all the examined HPE models,
reducing
15.1\% and 18.9\% error for in-domain and out-of-domain settings, respectively.
Code is available at~\url{https://github.com/Shimmer93/EMDUL}.
\end{abstract}    
\section{Introduction}
\label{sec:intro}
Human pose estimation (HPE) is to predict the human skeleton in terms of the locations and connectivity of the joints (or keypoints).  It has broad applications in robotics, human-computer interaction, action recognition, etc. 
Millimeter-wave (mmWave) 
HPE has drawn wide and sustained attention in recent years due to the strengths over traditional RGB cameras in terms of its 3D nature, user privacy protection, robustness against lighting conditions, etc. For training and inference purposes, mainstream mmWave HPE has adopted point cloud (PC)
due to its representation simplicity and processing efficiency.

\begin{figure}[t]
    \centering
    \begin{subfigure}[t]{.45\linewidth}
        \centering
        \includegraphics[width=\linewidth]{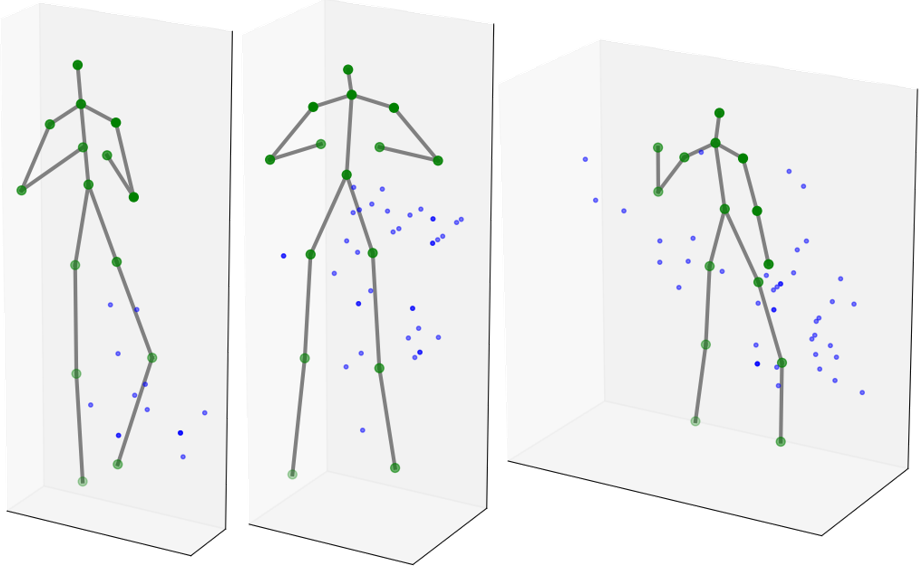}
        \caption{mmWave Samples}
    \end{subfigure}
    \begin{subfigure}[t]{.4\linewidth}
        \centering
        \includegraphics[width=\linewidth]{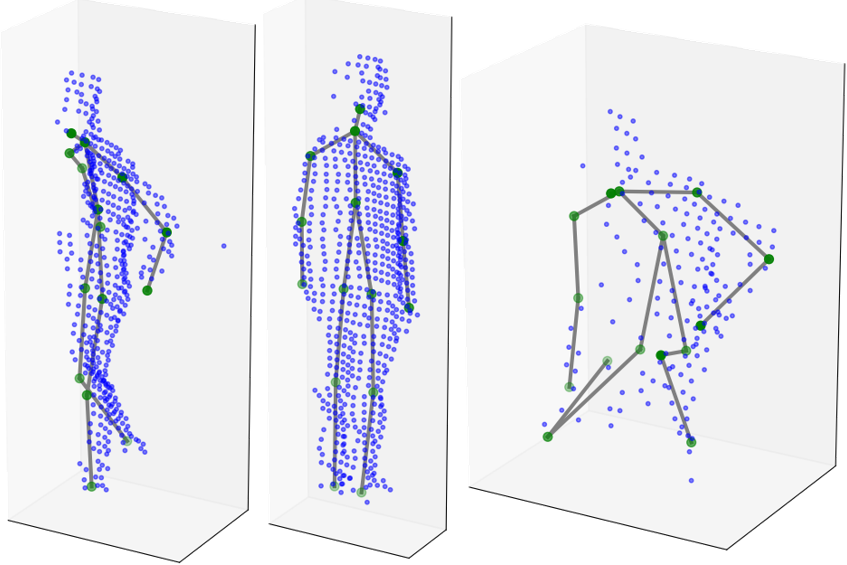}
        \caption{LiDAR Samples}
    \end{subfigure}
    \begin{subfigure}[t]{.205\linewidth}
        \centering
        \includegraphics[width=\linewidth]{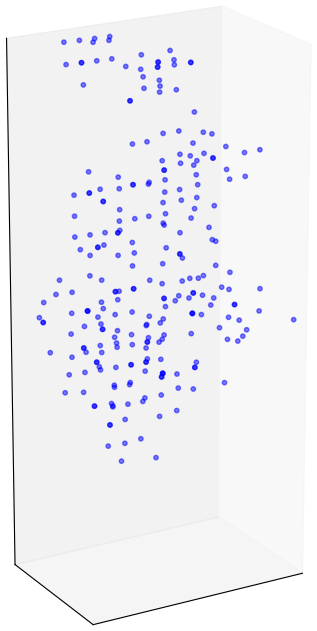}
        \caption{Unseen PC}
    \end{subfigure}
    \begin{subfigure}[t]{.205\linewidth}
        \centering
        \includegraphics[width=\linewidth]{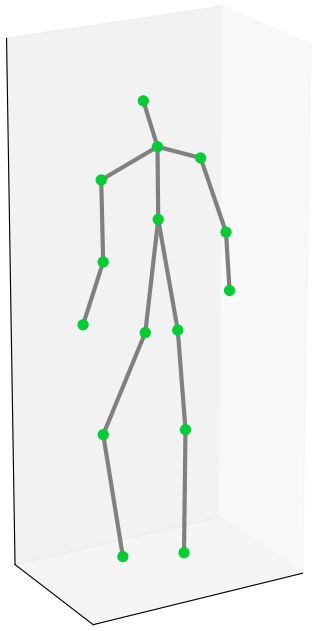}
        \caption{GT}
    \end{subfigure}
    \begin{subfigure}[t]{.205\linewidth}
        \centering
        \includegraphics[width=\linewidth]{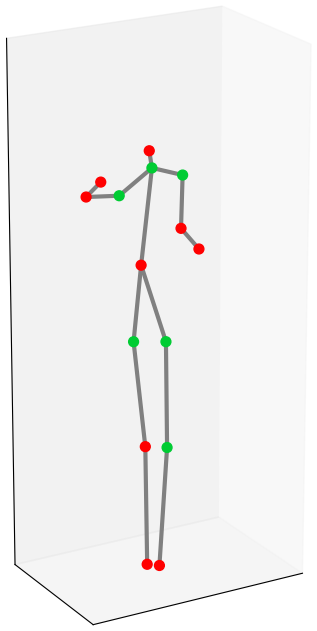}
        \caption{P4T}
    \end{subfigure}
    \begin{subfigure}[t]{.205\linewidth}
        \centering
        \includegraphics[width=\linewidth]{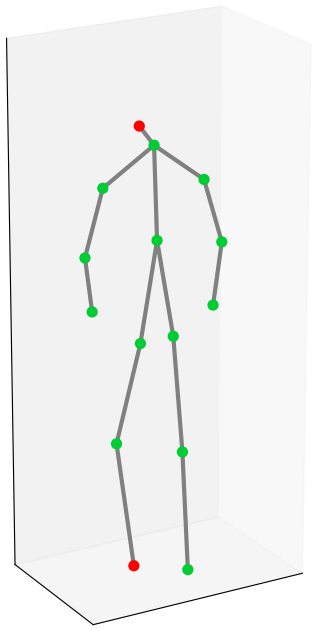}
        \caption{Ours}
    \end{subfigure}

    \caption{Examples illustrating the effect of dataset expansion. (a) Samples from an mmWave HPE training dataset. (b) Samples from a LiDAR dataset with richer pose diversity used for dataset expansion; (c) An mmWave PC from an unseen scenario. (d) The ground-truth skeleton. (e) The predicted skeleton of SOTA P4T~\cite{fanPoint4DTransformer2021} without expansion. (f) The predicted skeleton of P4T trained on \schname-expanded dataset. Joints are colored red for errors $>10\,\text{cm}$ and green otherwise. \schname\ achieves stronger generalization ability than the baseline P4T. }
    \label{fig:first}
\end{figure}

Despite the long-standing community interest in mmWave HPE, at present its labeled sets of data, the so-called ``datasets," are still scarce. Furthermore, their diversity is limited in two critical aspects: (1) \textit{PC attributes}, such as detection noise, point density, and 
human motion sensitivity, mainly due to lack of device heterogeneity and environmental variety during data collection;
and (2) \textit{Human poses}, where subjects often move in
simple 
postures facing the radar with marked uniformity. Such datasets severely undermine model generalization, leading to degraded performance in real-world or unseen environments. 

While labeled mmWave data is scarce, {\em unlabeled} mmWave data with diverse poses can be easily collected with minimal setup and
manual annotations. The key issue is how to annotate automatically, i.e., ``pseudo-label'', such unlabeled data to embrace them into the current mmWave datasets to enhance the data diversity for training.

We further note that the PC datasets of LiDAR, another prevalent sensor for HPE, are abundant and widely available, often covering more diverse poses than mmWave datasets.  
It would be desirable to leverage these LiDAR datasets for mmWave training. 
However, this goal is challenging to achieve because the intrinsic PC attributes of LiDAR are fundamentally different from those of mmWave (due to distinct physical principles of sensing).
We need to bridge the gap between the two modalities to make the LiDAR datasets useful for mmWave training.


To expand the existing mmWave datasets, we investigate, for the first time, how to {\em pseudo-label} the unlabeled mmWave PCs, and {\em convert (or translate)} a LiDAR dataset into its mmWave counterpart. 
Both the pseudo-labeled and converted mmWave data then  
greatly expand the original mmWave dataset for
mmWave HPE model training. 

Previous attempts to expand mmWave datasets are based on 
extracting and converting the skeletons from 
video datasets to mmWave PCs~\cite{dengMidasGeneratingMmWave2023, fanVideo2mmPointSynthesizingMmWave2025}. 
However, these methods focus on expanding human actions, whereas the PCs follow a similar distribution as the original mmWave data without diversifying their attributes.
Some other methods augment mmWave PCs using LiDAR PCs as supervision signals~\cite{chengNovelRadarPoint2022, prabhakaraHighResolutionPoint2023, hanDenserRadar4DMillimeterWave2024, luanDiffusionBasedPointCloud2024}.  While commendable, they require mmWave-LiDAR pairing and their simultaneous joint labeling in the dataset. Such multi-modal data is scarce, inconvenient to collect, and costly to label in practice, hence limiting their deployability. 

In this work, we make the following contributions:
\begin{itemize}

 \item{\em A Trained Pseudo-label Estimator for the Unlabeled mmWave Point Cloud:} 
 %
%
%
    To turn unlabeled mmWave PCs into training data, we propose a simple yet effective pseudo-label estimator trained with 
    a supervised loss on an mmWave dataset, and an unsupervised temporal consistency loss (UTCL) on unlabeled mmWave data.  Our UTCL improves the estimation robustness by enforcing temporal consistency in the predicted pseudo-labels.

    
    \item{\em A Closed-Form Converter to Translate LiDAR Dataset to mmWave Point Cloud:} We propose  a closed-form PC converter, an approach independent of pseudo-label estimator that translates a LiDAR dataset to its mmWave counterpart.
   The converter uses a flow-based point filtering (FPF) algorithm
    to realistically 
    capture
    the motion detection mechanism of mmWave PCs where moving body parts (high-flow points) are more likely to be detected than static ones (low-flow points). By integrating FPF with traditional PC augmentation techniques (\eg, noise injection, random sampling),
    \schname\ effectively translates an input LiDAR dataset into its mmWave counterpart with realistic PC attributes.
    
  \item{\em  \schname, A Novel Pipeline to Expand an mmWave Dataset:} We propose \schname, a novel pipeline to greatly {\bf e}xpand an {\bf m}mWave {\bf d}ataset 
with {\bf u}nlabeled mmWave data and {\bf L}iDAR dataset.
 \schname\ first uses the closed-form converter to translate LiDAR dataset to its mmWave counterpart.  Using the converted and existing mmWave datasets, it subsequently trains the pseudo-label estimator.  The trained estimator is then used to annotate unlabeled mmWave PC. 
Expanded with the converted and pseudo-labeled data, 
the original mmWave dataset is greatly enhanced in terms of volume and diversity.
This dataset is then used to train the final HPE inference models.

\end{itemize}

    


To validate \schname, we conduct extensive experiments on commonly used mmWave datasets, MM-Fi~\cite{yangMMFiMultiModalNonIntrusive2023} and mmBody~\cite{chenMmBodyBenchmark3D2023} with portions designated as unlabeled data, and LiDAR datasets LiDARHuman26M~\cite{liLiDARCapLongrangeMarkerless2022} and HmPEAR~\cite{linHmPEARDatasetHuman2024}.
Our results show that \schname\ significantly improves model performance and generalization compared to training solely on the original mmWave dataset. 
In our generalization study, \schname\ achieves a substantial 15.1\% error reduction when trained and tested on MM-Fi (in-domain study), 
and a 18.9\% reduction when trained on mmBody and tested on MM-Fi (out-of-domain study) using unlabeled data and HmPEAR for expansion.
\section{Related Works}

\subsection{mmWave-based Human Pose Estimation}
In recent years, mmWave radar has attracted increasing attention for HPE due to its low deployment cost, 3D data capture, privacy-friendly nature, and robustness under various lighting conditions. While some methods directly use raw 4D radar tensor as input~\cite{dingRadarBased3DHuman2021, leeHuPRBenchmarkHuman2023b, zhaoCubeLearnEndtoEndLearning2023, zhuProbRadarM3FMmWaveRadar2024, hoRTPose4DRadar2025}, the high computational cost and unavailability on certain hardware platforms limit their application scenarios. Therefore, mainstream mmWave HPE approaches often use processed 3D point clouds (PCs)~\cite{singhRadHARHumanActivity2019, anFastScalableHuman2022, cuiRealTimeShortRangeHuman2022, chenMmBodyBenchmark3D2023, senguptaMmPoseNLPNaturalLanguage2023, fanDiffusionModelGood2024}, which reduces the computational demand and ensures broader device compatibility.
However, existing datasets for mmWave PC-based HPE~\cite{anMRIMultimodal3D2022,chenMmBodyBenchmark3D2023,yangMMFiMultiModalNonIntrusive2023,cui2023milipoint} are scarce and lack diversity in PC attributes and human poses, leading to poor generalization on unseen testing scenarios. 


\begin{figure*}[t]
    \centering
    \includegraphics[width=\linewidth]{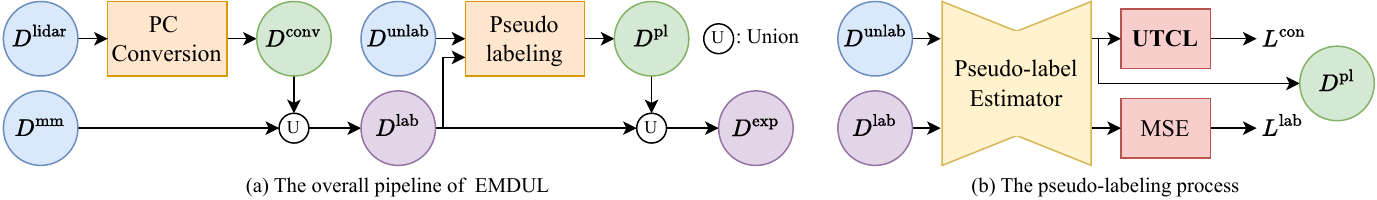}
    \caption{The overview of \schname{} integrating both PC conversion and pseudo-labeling modules. }
    \label{fig:lemt}
\end{figure*}

\subsection{Data Expansion or Augmentation for mmWave 
Datasets
}
To address the data scarcity issue in mmWave-based HPE, various data expansion or augmentation strategies have been investigated.
Current data {\em expansion} methods for HPE or related human sensing tasks are typically 
based on extracting additional skeletons from video data and converting them to PCs by leveraging statistical properties of the original mmWave dataset~\cite{dengMidasGeneratingMmWave2023, fanVideo2mmPointSynthesizingMmWave2025}. 
Although these methods increase pose diversity, the generated PCs still follow a similar distribution to the original dataset. 
Compared to mmWave, LiDAR HPE datasets also contain 3D PCs and generally capture a wider range of human poses \cite{liLiDARCapLongrangeMarkerless2022, linHmPEARDatasetHuman2024,renLiveHPSLiDARBasedSceneLevel2024}, making them an ideal resource for expanding mmWave data in both PC and pose diversity. 
However, due to the distinct attributes between LiDAR and mmWave PCs, models trained exclusively on LiDAR data exhibit poor generalization when applied to mmWave data. 
Existing {\em augmentation} techniques developed for other tasks address this issue by using LiDAR PCs as a supervision signal to improve mmWave PC density~\cite{chengNovelRadarPoint2022, prabhakaraHighResolutionPoint2023, hanDenserRadar4DMillimeterWave2024, luanDiffusionBasedPointCloud2024}. While effective, these methods rely on co-labeled mmWave-LiDAR pairs, which are challenging to acquire in practice. 
Our proposed method, in contrast, efficiently leverages an independent LiDAR dataset to diversify both PCs and human poses without requiring co-labeled data.

\subsection{Semi-Supervised Learning Approaches
}
Semi-supervised learning trains a model using limited labeled data alongside a large amount of unlabeled data. Most existing semi-supervised learning techniques fall into two categories: (1) \textit{pseudo-labeling-based methods}~\cite{leePseudoLabelSimpleEfficient2013, radosavovicDataDistillationOmniSupervised2018, xieSelfTrainingNoisyStudent2020}, which predict pseudo-labels for unlabeled data, and (2) \textit{consistency-based methods}~\cite{laineTemporalEnsemblingSemiSupervised2016, sajjadiRegularizationStochasticTransformations2016, tarvainenMeanTeachersAre2017, berthelotMixMatchHolisticApproach2019, sohnFixMatchSimplifyingSemisupervised2020}, which apply different augmentations to the same input and enforce output similarity. Based on these techniques, various methods have been proposed for image-based HPE~\cite{pavllo3DHumanPose2019, xieEmpiricalStudyCollapsing2021, huangSemiSupervised2DHuman2023}, but they cannot be directly applied to mmWave HPE due to differences in input modality and output formats. Our \schname\ incorporates a pseudo-labeling method specifically designed for mmWave HPE, effectively utilizing unlabeled mmWave data for dataset expansion.  

\section{Problem Formulation and \schname{} Overview} \label{sec:problem}

\subsection{Problem Formulation} \label{sec:basic}
Let $D$ denote an HPE dataset comprising one or multiple sequences of PC-skeleton pairs $(P,S)$ with variable lengths. 
At each timestep $t\geq 0$, the input for an HPE model is a sequence of $T$ continuous PCs $\{P_{t-T+1}, ..., P_t\}$. Each PC $P_{j}\in \mathbb{R}^{M_j\times 3}$ consists of $M_j$ points represented by their Cartesian coordinates. The corresponding output is the human joint coordinates $\hat{S}_t\in\mathbb{R}^{J\times 3}$ at time $t$, where $J$ is the number of joints in the skeleton. Our objective is to expand an mmWave dataset $D^{\text{mm}}$ in terms of volume and diversity.

While mmWave PCs can include features like Doppler speed, these are often hardware-dependent and can limit a model's generalization ability.
We therefore exclude them. 
For HPE models that require a per-point feature (\eg P4Transformer~\cite{fanPoint4DTransformer2021}), we use only the point's height, which is independent of specific hardware.

\subsection{\schname{} Overview}

We propose \schname{}, a novel approach to expand an mmWave HPE dataset $D^{\text{mm}}$ using two readily available data sources: unlabeled mmWave data $D^{\text{unlab}}$ and an annotated LiDAR dataset $D^{\text{lidar}}$. 
\schname{} consists of two independent modules: 
\begin{itemize}
\item \textit{Pseudo-labeling of unlabeled data}, which generates pseudo-labels for $D^{\text{unlab}}$ to form $D^{\text{pl}}$.
\item \textit{PC conversion of LiDAR datasets}, which translates PCs in $D^{\text{lidar}}$ into its mmWave counterpart to form $D^{\text{conv}}$. 
\end{itemize}
The complete pipeline of \schname\ is displayed in~\cref{fig:lemt}(a).
First, $D^{\text{lidar}}$ is translated into $D^{\text{conv}}$ using the PC conversion module, which is then combined with $D^{\text{mm}}$ to form $D^{\text{lab}}=D^{\text{mm}}\cup D^{\text{conv}}$. 
Next, the pseudo-labeling module processes $D^{\text{lab}}$ and $D^{\text{unlab}}$ to generate pseudo-labels for $D^{\text{unlab}}$, yielding $D^{\text{pl}}$. The inclusion of $D^{\text{conv}}$ in the training data substantially improves the quality of generated pseudo-labels. 
The final expanded dataset is the union: $D^{\text{exp}}=D^{\text{lab}}\cup D^{\text{pl}}$.

\subsection{Training on Expanded Dataset}
The inference HPE model $\theta^{\text{infer}}$ is trained from scratch on the expanded dataset $D^{\text{exp}}$ with a mean-squared error (MSE) loss. 
To maximize diversity, $D^{\text{conv}}$ is re-generated at each epoch with a new random seed.
Concurrently, the pseudo-estimator $\theta^{\text{pl}}$ is trained alongside $\theta^{\text{infer}}$. During each epoch, $\theta^{\text{pl}}$ is updated first and used to generate a new $D^{\text{pl}}$ from $D^{\text{unlab}}$. $\theta^{\text{infer}}$ is then trained on the updated $D^{\text{exp}}=D^{\text{lab}}\cup D^{\text{pl}}$. This iterative refinement allows $\theta^{\text{infer}}$ to incrementally extract meaningful information from $D^{\text{unlab}}$.

\section{Pseudo-labeling of Unlabeled mmWave Data} \label{sec:japl}

To utilize unlabeled mmWave data for dataset expansion, we employ a simple but effective pseudo-labeling estimator.
As illustrated in~\cref{fig:lemt}(b), the estimator $\theta^{\text{pl}}$ is trained on labeled data $D^{\text{lab}}$ using an MSE loss $L^{\text{lab}}$, as well as simultaneously on unlabeled data $D^{\text{unlab}}$ using our novel Unsupervised Temporal Consistency Loss (UTCL) $L^{\text{con}}$ (\cref{sec:ucl}). The $\theta^{\text{pl}}$ generates pseudo-labels for $D^{\text{unlab}}$ to form a new dataset $D^{\text{pl}}$.





\subsection{Unsupervised Temporal Consistency Loss (UTCL)} \label{sec:ucl}
\begin{figure}[t]
    \centering
    \includegraphics[width=0.3\linewidth]{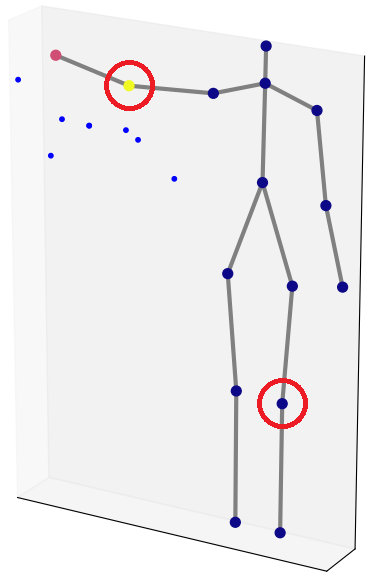}
    \caption{Illustration of the motion-detection mechanism in mmWave radar using an MM-Fi sample. Joints with high flow (yellow) lie close to detected points, while low-flow joints (dark blue) have no nearby points.}
    \label{fig:ucl}
\end{figure}

Predicting skeletons independently at each timestep can lead to temporal inconsistency. To enhance pseudo-label reliability, we propose UTCL, motivated by a key physical insight from mmWave sensing: joints far from any detected points are likely static, whereas those embedded within the PC are likely in motion.
This arises from the \textit{motion detection mechanism} (\cref{fig:ucl}) of mmWave radar: due to the reliance on the Doppler effect during PC formation, mmWave radars detect moving targets more easily than static ones. 
UTCL enforces this physical prior by penalizing predictions that violate it. It consists of two complementary components: a Dynamic Consistency Loss (DCL) and a Static Consistency Loss (SCL).

Given a PC $P_t$, its predicted skeleton $\hat{S}_t$, and the skeleton flow $\hat{F}_t^S = \hat{S}_t - \hat{S}_{t-1}$, DCL encourages joints that are likely moving to have a non-zero flow. We first identify the set of ``dynamic'' joints, $F_t^{\text{dyn}}$, as those whose distance to the nearest point in $P_t$ is less than a threshold $\mu$:
\begin{align}
    F_t^{\text{dyn}} &= \{\hat{F}^{S}_{t}[j]:\min_i{\lVert \hat{S}_{t}[j] -  P_{t}[i]\rVert_2} < \mu\}.
\end{align}
Then, $L^{\text{dyn}}$ penalizes these joints if their flow magnitude is below a flow threshold $\eta$, effectively encouraging motion:
\begin{align}
    L^{\text{dyn}} &= \frac{1}{\lvert F_t^{\text{dyn}}\rvert}\sum_{k=1}^{\lvert F_t^{\text{dyn}}\rvert}\max{(0, \eta - \lVert F_{t}^{\text{dyn}}[k] \rVert_2 )}.
\end{align}

Similarly, SCL encourages joints that are likely static to have zero flow:
\begin{align}
    F_t^{\text{sta}} &= \{\hat{F}^{S}_{t}[j]:\min_i{\lVert \hat{S}_{t}[j] -  P_{t}[i]\rVert}_2 > \rho\},\\
    L^{\text{sta}} &= \frac{1}{\lvert F_t^{\text{sta}}\rvert}\sum_{k=1}^{\lvert F_t^{\text{sta}}\rvert}\lVert F_{t}^{\text{sta}}[k] \rVert_2,
\end{align}
where $\rho$ is the threshold for static joints.

The final UTCL is the sum of DCL and SCL: $L^{\text{con}} = L^{\text{dyn}} + L^{\text{sta}}$.

\subsection{Training of Pseudo-label Estimator}
In each training step, we sample one instance from $D^{\text{lab}}$ and another from $D^{\text{unlab}}$.
The overall training loss combines the supervised loss $L^{\text{lab}}$ and UTCL $L^{\text{con}}$ as a weighted sum:
\begin{align}
    L = L^{\text{lab}} + \lambda^{\text{con}} L^{\text{con}},
\end{align}
where $\lambda^{\text{con}}$ is the weighting parameter.

\section{Converting LiDAR Datasets to mmWave Point Clouds}\label{sec:convert}
In addition to pseudo-labeling unlabeled mmWave data, we also convert a LiDAR dataset into its mmWave counterpart to expand the original mmWave dataset. 

\subsection{Closed-Form PC Conversion}
Our closed-form PC conversion pipeline is a sequence of augmentations designed to simulate different attributes of mmWave PCs:

\begin{itemize}
    \item \textit{Noisy point addition (NPA)}: Adds a fixed number of noise points to simulate non-human environmental objects detected by an mmWave radar.
    \item \textit{Flow-based Point Filtering (FPF)}: Our proposed algorithm (detailed in~\cref{sec:fpf}) that simulates the motion detection mechanism as described in~\cref{sec:ucl}.
    \item \textit{Random sampling (RS)}: Reduces point density by randomly sampling a fraction of points, mimicking sparser mmWave PCs.
    \item \textit{Noise injection (NI)}: Injects random noise into each point's coordinates to simulate lower spatial resolution.
\end{itemize}

\begin{figure}[t]
    \centering
    \begin{subfigure}[t]{.19\linewidth}
        \centering
        \includegraphics[width=\linewidth]{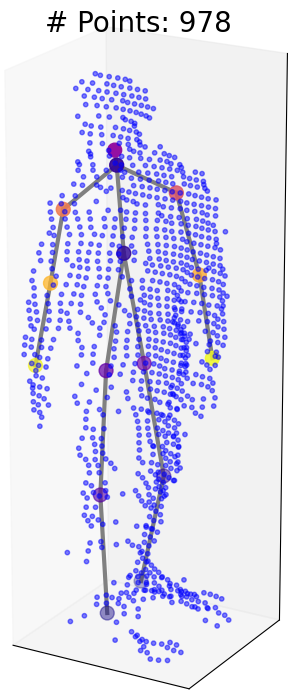}
        \caption{Original}
    \end{subfigure}
    \begin{subfigure}[t]{.19\linewidth}
        \centering
        \includegraphics[width=\linewidth]{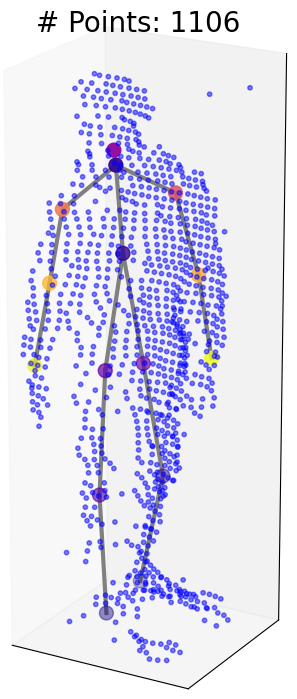}
        \caption{After NPA}
    \end{subfigure}
    \begin{subfigure}[t]{.19\linewidth}
        \centering
        \includegraphics[width=\linewidth]{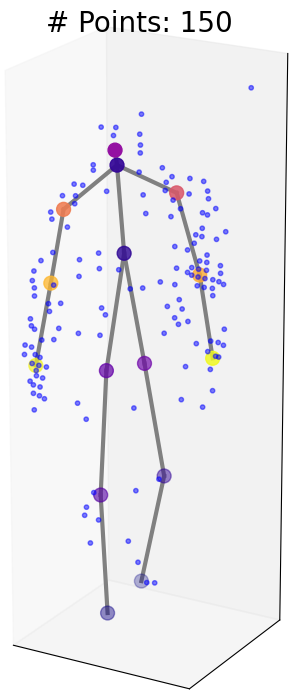}
        \caption{After FPF}
    \end{subfigure}
    \begin{subfigure}[t]{.19\linewidth}
        \centering
        \includegraphics[width=\linewidth]{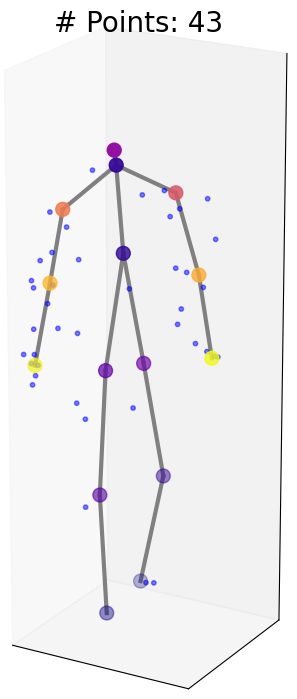}
        \caption{After RS}
    \end{subfigure}
    \begin{subfigure}[t]{.19\linewidth}
        \centering
        \includegraphics[width=\linewidth]{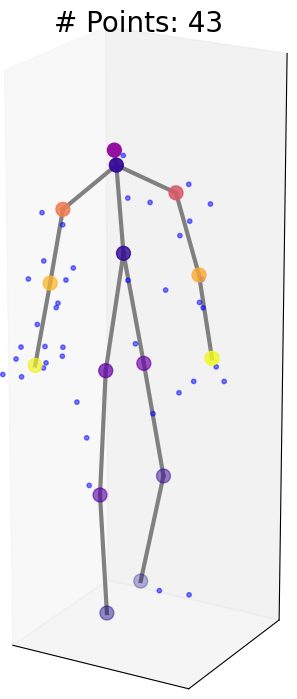}
        \caption{After NI}
    \end{subfigure}

    \caption{Step-by-step visualization of the point-cloud (PC) conversion pipeline
    Blue joints have lower flow magnitudes and yellow joints higher ones.}
    \label{fig:fpf}
\end{figure}

The augmentations are sequentially applied in the order: $\text{NPA}\rightarrow{}\text{FPF}\rightarrow{}\text{RS}\rightarrow{}\text{NI}$. The progressive transformation after each step is visualized in~\cref{fig:fpf}, which clearly illustrates their impacts on PC attributes.

\subsection{Flow-based Point Filtering (FPF)} \label{sec:fpf}
Flow refers to the temporal displacement of points or joints. FPF simulates the motion detection mechanism of mmWave radar by interpolating PC flow based on skeleton flow, and then filtering points with lower flow magnitude with higher probability. 

Given two consecutive LiDAR PCs, $P_{t-1}$ and $P_t$, and their ground-truth skeletons, $S_{t-1}$ and $S_t$, we aim to estimate a 3D flow vector $F_t^P[i]$ for each point $P_t[i]$ in the current PC.
To create a stable flow field for interpolation, we establish boundary constraints by extending the skeletons $S_{t-1}$ and $S_t$ to $S'_{t-1}$ and $S'_{t}$ by adding the eight vertices of the axis-aligned bounding cube that encloses both the joints and PCs at $t-1$ and $t$. We define the extended skeleton flow $F_{t}'^{S}$ as the displacement of these $J+8$ points $S'_t - S'_{t-1}$. 
Notably, the eight static vertices naturally have zero flow.

We then interpolate the flow for each point $P_t[i]$ as a linear combination of $F_{t}'^{S}$ using inverse distance weighting. The normalized weights $\tilde{w}_t[i,j]$ for each point $i$ relative to each of the $J+8$ extended joints $j$ is:
\begin{align}
    w_{t}[i,j] &= \frac{1}{\lVert P_{t}[i] - S'_{t}[j] \rVert_2+\epsilon},\\
    \tilde{w}_{t}[i,j] &= \frac{w_{t}[i,j]}{\sum_{k=1}^{J+8} w_{t}[i,k]},
\end{align}
where $\epsilon=10^{-6}$ is used to prevent division by zero.
The interpolated PC flow $F_t^P$ is then:
\begin{align}
    F^P_{t} = \tilde{w}_t F'^S_t.
\end{align}
This ensures that the flow of each point is most similar to its nearest joints or boundary vertices.


Finally, we simulate the motion detection mechanism by filtering $P_t$ using the interpolated flow $F_t^P$. Points in $P_{t}$ with lower flow magnitude are discarded with a higher probability to form the resultant PC $P^{\text{conv}}_{t}$:
\begin{align}
    \mathcal{P}(P_{t}[i]\in P^{\text{conv}}_{t}) = \min{\left(\frac{\lVert F^P_{t}[i]\rVert_2}{\upsilon_t}, 1\right)},
\end{align}
where $\upsilon_t$ is the flow threshold sampled from a uniform distribution $\text{U}[\gamma, \delta]$, with $\gamma$ and $\delta$ as hyperparameters. 

\begin{figure}[t]
    \centering
    \includegraphics[width=\linewidth]{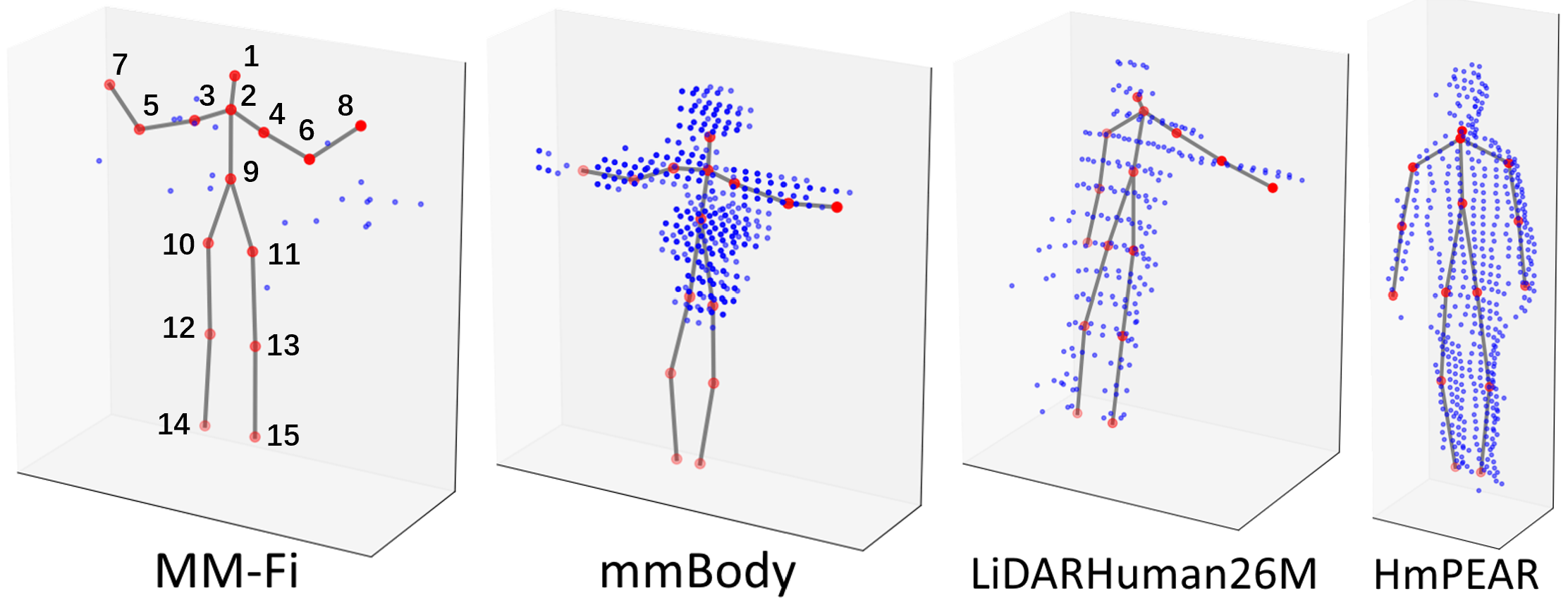}
    \caption{Sample point clouds from different mmWave and LiDAR HPE datasets and the standardized 15-keypoint skeleton structure used in this paper. 
    }
    \label{fig:skl}
\end{figure}

\begin{table*}[t]
\centering
\small
\caption{
Comparison with state-of-the-art mmWave HPE methods.
Each model is trained with only 10\% labeled data from MM-Fi (F) or mmBody (B).
Depending on the setting, methods may additionally use the remaining 90\% unlabeled mmWave data and/or the LiDAR dataset HmPEAR.
All results are reported in centimeters (cm), and lower is better.
}
\resizebox{\linewidth}{!}{
\begin{tabular}{ll|cc|cc|cc|cc}
\toprule
\multirow{2}{*}{Method} & \multirow{2}{*}{Backbone}
& \multicolumn{2}{c|}{F $\rightarrow$ F}
& \multicolumn{2}{c|}{F $\rightarrow$ B}
& \multicolumn{2}{c|}{B $\rightarrow$ F}
& \multicolumn{2}{c}{B $\rightarrow$ B} \\
& & MPJPE & PA-MPJPE & MPJPE & PA-MPJPE & MPJPE & PA-MPJPE & MPJPE & PA-MPJPE \\
\midrule

\multicolumn{10}{c}{\textbf{10\% labeled mmWave only}} \\
\midrule
PT~\cite{zheng3DHumanPose2021}         & PT   & 15.88 & 10.62 & \textbf{17.75} & \textbf{14.03} & 35.22 & \textbf{14.83} & 13.32 &  9.78 \\
mmDiff~\cite{fanDiffusionModelGood2024} & mmDiff & 15.10 & 10.73 & 44.00 & 21.76 & \textbf{28.12} & 20.15 & 14.98 & 10.69 \\
P4T~\cite{fanPoint4DTransformer2021}    & P4T   & 12.23 &  7.95 & 20.78 & 16.10 & 33.62 & 15.85 & 11.39 &  8.37 \\
SPiKE~\cite{ballesterSPiKE3DHuman2024}  & SPiKE & \textbf{11.85} & \textbf{7.92} & 18.70 & 14.40 & 37.09 & 16.30 & \textbf{10.70} & \textbf{7.86} \\
\midrule

\multicolumn{10}{c}{\textbf{+ unlabeled mmWave data}} \\
\midrule
MT~\cite{tarvainenMeanTeachersAre2017}           & P4T   & 26.77 & 14.94 & \textbf{19.27} & \textbf{15.27} & 52.52 & 21.68 & 15.43 & 10.00 \\
MT~\cite{tarvainenMeanTeachersAre2017}           & SPiKE & 19.62 & 11.03 & 20.56 & 16.21 & 47.82 & 17.50 & 12.89 &  8.68 \\
\textbf{\schname-PL} & P4T   & \textbf{11.36} &  7.38 & 22.90 & 16.13 & 41.47 & 17.98 & \textbf{10.79} &  8.26 \\
\textbf{\schname-PL} & SPiKE & 11.45 & \textbf{7.21} & 23.93 & 16.04 & \textbf{38.67} & \textbf{17.01} & 10.83 & \textbf{6.99} \\
\midrule

\multicolumn{10}{c}{\textbf{+ HmPEAR LiDAR data}} \\
\midrule
P4T~\cite{fanPoint4DTransformer2021}    & P4T   & 11.02 &  7.55 & 16.08 & 12.93 & 32.13 & 14.93 & 11.14 &  7.17 \\
SPiKE~\cite{ballesterSPiKE3DHuman2024}  & SPiKE & 10.41 &  7.15 & 17.29 & 12.81 & 33.90 & 15.34 & 10.90 & \textbf{6.86} \\
\textbf{\schname-PCC} & P4T   & \textbf{10.21} & \textbf{7.12} & \textbf{15.22} & \textbf{11.51} & \textbf{24.24} & 14.99 & \textbf{10.86} &  7.09 \\
\textbf{\schname-PCC} & SPiKE & 10.59 &  7.36 & 15.41 & 11.71 & 24.25 & \textbf{14.59} & 10.97 &  7.11 \\
\midrule

\multicolumn{10}{c}{\textbf{Full setting (+ unlabeled mmWave data + HmPEAR LiDAR data)}} \\
\midrule
MT~\cite{tarvainenMeanTeachersAre2017}           & P4T   & 10.37 & \textbf{6.80} & 16.97 & 12.12 & 31.51 & 14.61 & 11.04 &  7.17 \\
MT~\cite{tarvainenMeanTeachersAre2017}           & SPiKE & 12.45 &  7.54 & 16.15 & 12.15 & 32.71 & 15.74 & 11.04 &  7.17 \\
\textbf{\schname}    & P4T   & \textbf{10.06} &  7.01 & \textbf{14.89} & \textbf{11.11} & 24.01 & 14.33 & 11.11 & \textbf{6.89} \\
\textbf{\schname}    & SPiKE & 10.40 &  7.23 & 15.11 & 11.36 & \textbf{22.80} & \textbf{14.09} & \textbf{10.82} &  7.02 \\
\bottomrule
\end{tabular}}
\label{tab:exp_main}
\end{table*}

    

\section{Illustrative Experimental Results}
In this section, the datasets used in experiments are first introduced in ~\cref{sec:data}. We then detail the implementation in ~\cref{sec:implement} and evaluation metrics in ~\cref{sec:metric}. Next, the experimental results compared with various HPE methods are shown in ~\cref{sec:comp}. Finally, ~\cref{sec:ablation} presents ablation studies and additional analysis on \schname.

\subsection{Datasets}\label{sec:data}
We evaluate \schname\ on two mmWave datasets:

\begin{itemize}
    \item{\em mmBody (B)}~\cite{chenMmBodyBenchmark3D2023} is a multi-modal human sensing dataset incorporating a high-end mmWave radar, offering denser PCs that capture more static body parts than standard radars. It covers 9 scenes with 200K frames. Although it includes 200 motions performed by 30 subjects, most only involve upright postures facing the radar, resulting in limited pose diversity. 
    
    \item{\em MM-Fi (F)}~\cite{yangMMFiMultiModalNonIntrusive2023} is a comprehensive human pose dataset that includes mmWave radar as one modality. Compared with mmBody, PCs in MM-Fi are sparser and less sensitive to static body parts. It contains 321K frames across 4 scenes, covering 27 actions by 40 subjects with similarly limited pose diversity. 
\end{itemize}

We further use two LiDAR datasets:

\begin{itemize}
    \item{\em LiDARHuman26M (L)}~\cite{liLiDARCapLongrangeMarkerless2022} is a long-range LiDAR-based human pose dataset capturing human actions at various distances, resulting in PCs with varying, sometimes low, densities. It contains 184K frames across 2 scenes and 20 actions with more diverse poses, including ones unseen in mmBody and MM-Fi (e.g., swimming and squatting).

    \item{\em HmPEAR (H)}~\cite{linHmPEARDatasetHuman2024} is a recent LiDAR dataset for HPE and action recognition, consisting of 250K frames from 10 scenes. It includes 40 action categories with even greater pose diversity than LiDARHuman26M.
\end{itemize}

To unify skeleton formats, we adopt a standardized structure with 15 keypoints (\cref{fig:skl}). To simulate extreme mmWave data scarcity, $D^{\text{mm}}$ contains PCs and skeletons from 10\% randomly sampled sequences of the mmWave training set, while PCs from the remaining 90\% form the unlabeled dataset $D^{\text{unlab}}$. 
$D^{\text{lidar}}$ includes the full training set of the LiDAR dataset.

\subsection{Implementation}\label{sec:implement}
In pseudo-labeling, we use P4T~\cite{fanPoint4DTransformer2021} or SPiKE~\cite{ballesterSPiKE3DHuman2024} as the pseudo-label estimator. Its input is a sequence of 5 PCs, each uniformly processed to 256 points through truncation or padding. The thresholds in UTCL are configured as $\mu = 20\,\text{cm}$, $\eta = 5\,\text{cm}$ and $\rho = 5\,\text{cm}$, with weighting parameter
$\lambda^{\text{con}} = 0.01$. 
We train the estimator for 100 epochs using the AdamW~\cite{loshchilovDecoupledWeightDecay2017} optimizer with a learning rate $10^{-4}$, and the Cosine Annealing learning rate scheduler~\cite{loshchilovSGDRStochasticGradient2017} with linear warmup at an initial learning rate $10^{-5}$. 
In FPF, the flow threshold $\upsilon$ for FPF is sampled uniformly between $\gamma=2\,\text{cm}$ and $\delta=5\,\text{cm}$. 
The inference HPE model mirrors the estimator's network architecture, input format, and optimization configuration. 
Additional implementation details are provided in the supplementary material.

\subsection{Evaluation Metrics}\label{sec:metric} \
Following previous works, we employ two commonly used evaluation metrics from Human36M~\cite{ionescuHuman36MLargeScale2014}: 
\begin{itemize}
    \item \textit{Mean Per Joint Position Error (MPJPE)}, which calculates the average Euclidean distance error for each joint between the predicted skeleton and the ground truth.
    \item \textit{Procrustes Analysis MPJPE (PA-MPJPE)}, which measures error after aligning the predicted and ground-truth skeletons using Procrustes methods, including translation, rotation, and scaling, evaluating the quality of the overall pose structure.
\end{itemize}

\subsection{Comparison with the State of the Art}\label{sec:comp}
This section presents quantitative results of mmWave HPE models trained on \schname-expanded dataset compared to those with varying data conditions. In addition to evaluating our entire approach, we separately assess our pseudo-labeling for unlabeled data (\schname-PL) and PC conversion method for LiDAR data (\schname-PCC). 
We classify the comparison schemes into four categories based on the data used to expand $D^{\text{lab}}$: 

\begin{itemize}
    \item{\textit{No data expansion}}, where PT~\cite{zheng3DHumanPose2021}, mmDiff~\cite{fanDiffusionModelGood2024}, 
    P4T~\cite{fanPoint4DTransformer2021}, and SPiKE~\cite{ballesterSPiKE3DHuman2024} are trained solely on $D^{\text{mm}}$.
    \item{\textit{Expansion using a LiDAR dataset}}, where P4T and SPiKE are trained on both $D^{\text{mm}}$ and $D^{\text{lidar}}$.
    \item{\textit{Expansion using unlabeled data}}, where P4T and SPiKE are trained on $D^{\text{lab}}=D^{\text{mm}}$ and $D^{\text{unlab}}$, integrating a widely used Mean-Teacher (MT) pseudo-labeling strategy~\cite{tarvainenMeanTeachersAre2017} adapted by us for mmWave HPE. 
    \item{\textit{Expansion using both a LiDAR dataset and unlabeled data}}, where P4T and SPiKE are trained on $D^{\text{lab}}=D^{\text{mm}}\cup D^{\text{lidar}}$ and $D^{\text{unlab}}$ using MT for pseudo-labeling. 
\end{itemize}
Let $D^{\text{train}} \to D^{\text{test}}$ denote the setting where the model is trained on dataset $D^{\text{train}}$ and tested on $D^{\text{test}}$. The trained models are evaluated on both the test split of $D^{\text{mm}}$ (in-domain scenario, \eg, F $\to$ F) and another mmWave dataset with unseen scenarios (out-of-domain scenario, \eg, F $\to$ B). All results are presented in centimeters (cm).

The results presented in ~\cref{tab:exp_main} show that dataset expansion generally improves performance.
Specifically, pseudo-labeling (\schname-PL) of unlabeled data achieves better fitting for in-domain data; however, it also tends to overfit because $D^{\text{unlab}}$ and $D^{\text{mm}}$ share similar distributions in our setting. 
In contrast, PC conversion (\schname-PCC) for a LiDAR dataset substantially improves performance in both in-domain and out-of-domain scenarios.

When integrating both components, our proposed \schname\ consistently outperforms models with no or incomplete data expansion across most settings, achieving a significant 15.1\% decrease in MPJPE on F $\to$ F, and a 18.9\% MPJPE reduction on B $\to$ F. 
Compared to the MT approach, which also expands $D^{\text{mm}}$ with $D^{\text{lidar}}$ and $D^{\text{unlab}}$, \schname\ performs better in 7 out of 8 cases. When employing SPiKE as the same HPE model, \schname\ surpasses MT by 17.5\% in MPJPE on F $\to$ F, and 30.3\% in MPJPE on B $\to$ F. 
Results in~\cref{tab:exp_lidardata} show that the improvement is general across different LiDAR datasets. 
These validate that \schname\ significantly improves in-domain accuracy and out-of-domain generalization of models. Notably, the performance improvement is more pronounced in out-of-domain scenarios, indicating that \schname\ effectively enhances data diversity.

\begin{table}[t]
  \centering
  \small
  \caption{Comparison with Mean Teacher (MT) pseudo-labeling when expanding MM-Fi (F) with different LiDAR datasets. P4T~\cite{fanPoint4DTransformer2021} serves as the common HPE model.}
  \resizebox{\linewidth}{!}{
  \begin{tabular}{c|c|cc|cc}
    \toprule
    \multicolumn{2}{c|}{Setting} & \multicolumn{2}{c|}{F $\to$ F} & \multicolumn{2}{c}{F $\to$ B} \\
    \midrule
    Method & \makecell{LiDAR\\Dataset} & MPJPE & \makecell{PA-\\MPJPE} & MPJPE & \makecell{PA-\\MPJPE} \\
    \midrule
    MT       & H & 10.37 & \textbf{6.80} & 16.97 & 12.12 \\
    \schname & H & \textbf{10.06} &  7.01 & \textbf{14.89} & \textbf{11.11} \\
    \midrule 
    MT       & L   & 10.93 &  \textbf{6.85} & 18.62 & 16.31  \\
    \schname & L   & \textbf{10.05} &  6.93 & \textbf{16.97} & \textbf{12.82}  \\
    \midrule
    MT       & H+L   & 10.60 &  6.84 & 17.04 &  13.53 \\
    \schname & H+L   &  \textbf{9.92} &  \textbf{6.82} & \textbf{16.43} &  \textbf{12.34}  \\
  \bottomrule
  \end{tabular}}
  \label{tab:exp_lidardata}
\end{table}

\begin{table}[t]
  \centering
  \small
  \caption{Preliminary error comparison of different point features.}
  \begin{tabular}{c|cc}
    \toprule
    Point feature & MPJPE & PA-MPJPE \\
    \midrule
    Doppler & 25.15 & 16.25 \\
    Height  & \textbf{19.68} & \textbf{15.24} \\
    
  \bottomrule
  \end{tabular}
  \label{tab:feature}
\end{table}

\subsection{Ablation Studies}\label{sec:ablation}
In this section, we conduct ablation studies and other analyses under the setting of training on MM-Fi + HmPEAR and testing on MM-Fi or mmBody. All methods use P4T as the HPE model. Results are presented in centimeters (cm). 
We assess performance in out-of-domain evaluations (F $\to$ B) in ablation studies, as our primary goal is to improve model generalization rather than in-domain fitting.

\begin{table}[t]
    \centering
    \small
    \caption{Ablation study on pseudo-labeling of unlabeled data. 
    } 
    \begin{tabular}{ccc|cc}
      \toprule
      \multicolumn{3}{c|}{Components} & \multicolumn{2}{c}{F $\to$ B} \\
      \midrule
      $L^{\text{lab}}$ & $L^{\text{dyn}}$ & $L^{\text{sta}}$ & MPJPE &PA-MPJPE \\
      \midrule
                &           &           & 15.22 & 11.51 \\
      \ding{51} &           &           & 15.12 & 11.44 \\
      \ding{51} & \ding{51} &           & 15.35 & 11.73 \\
      \ding{51} &           & \ding{51} & 15.53 & 11.80 \\
      \midrule
      \ding{51} & \ding{51} & \ding{51} & \textbf{14.89} & \textbf{11.11} \\
      \bottomrule
    \end{tabular}
    \label{tab:pl}
\end{table}

\begin{table}[t]
    \centering
    \small
    \caption{Ablation study on PC conversion of a LiDAR dataset.
    } 
    \begin{tabular}{cccc|cc}
      \toprule
      \multicolumn{4}{c|}{Components} & \multicolumn{2}{c}{F $\to$ B} \\
      \midrule
      NPA & FPF & RS & NI & MPJPE & PA-MPJPE \\
      \midrule
                &           &           &           & 15.85 & 12.23 \\
      \ding{51} &           &           &           & 15.80 & 12.23 \\
                & \ding{51} &           &           & 15.63 & 11.80 \\
      \ding{51} & \ding{51} &           &           & 15.49 & 11.74 \\
      \ding{51} & \ding{51} & \ding{51} &           & 15.47 & 11.56 \\
      \ding{51} &           & \ding{51} & \ding{51} & 15.54 & 11.75 \\
      \midrule
      \ding{51} & \ding{51} & \ding{51} & \ding{51} & \textbf{14.89} & \textbf{11.11} \\
      \bottomrule
    \end{tabular}
    \label{tab:fpf}
\end{table}

\begin{table*}[t]
\centering
\small
\caption{Ablation study on hyperparameters in UTCL and FPF under F$\to$B.}
\begin{tabular}{ccc|ccc|ccc|ccc|ccc}
\toprule
\multicolumn{9}{c|}{UTCL Hyperparameters} & \multicolumn{6}{c}{FPF Hyperparameters}\\
\midrule
$\mu$ & MPJPE & \makecell{PA-\\MPJPE} & $\eta$ & MPJPE & \makecell{PA-\\MPJPE} &
$\rho$ & MPJPE & \makecell{PA-\\MPJPE} & $\gamma$ & MPJPE & \makecell{PA-\\MPJPE} &
$\delta$ & MPJPE & \makecell{PA-\\MPJPE} \\
\midrule
1 & 15.37 & 11.67 & 1 & 15.28 & 11.37 & 10 & 15.67 & 11.90 & 1 & 15.43 & 11.38 & 2 & 15.25 & 11.60 \\
5 & \textbf{14.89} & \textbf{11.11} & 5 & \textbf{14.89} & \textbf{11.11} & 20 & \textbf{14.89} & \textbf{11.11} & 2 & \textbf{14.89} & \textbf{11.11} & 5 & \textbf{14.89} & \textbf{11.11} \\
10 & 15.52 & 11.86 & 10 & 15.67 & 11.74 & 50 & 14.93 & 11.12 & 5 & 15.47 & 11.78 & 10 & 15.58 & 11.48 \\
\bottomrule
\end{tabular}
\label{tab:exp_hyper}
\end{table*}

\begin{figure}[t]
    \centering
    \begin{subfigure}[t]{.325\linewidth}
        \centering
        \includegraphics[width=\linewidth]{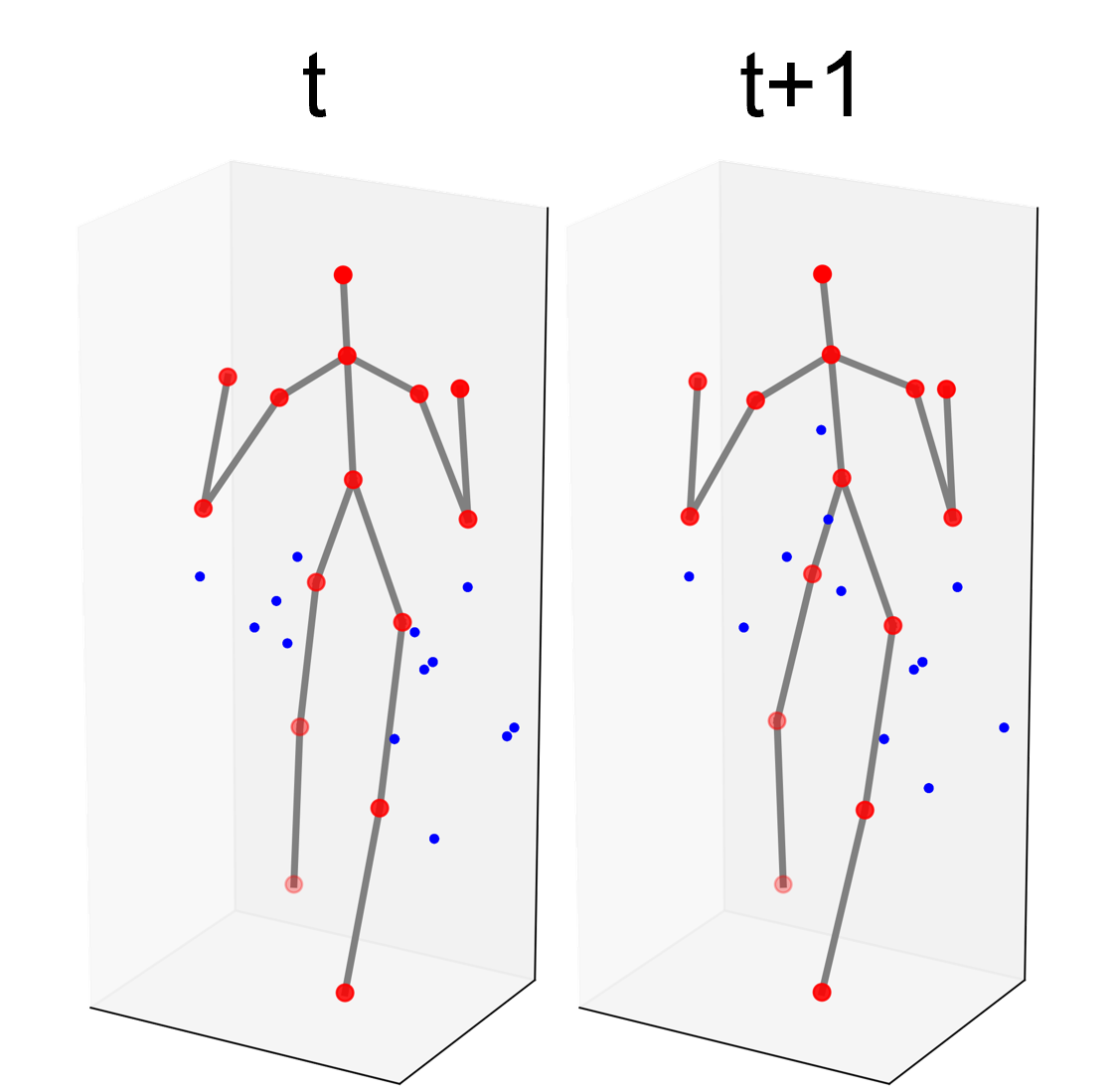}
        \caption{GT}
    \end{subfigure}
    \begin{subfigure}[t]{.325\linewidth}
        \centering
        \includegraphics[width=\linewidth]{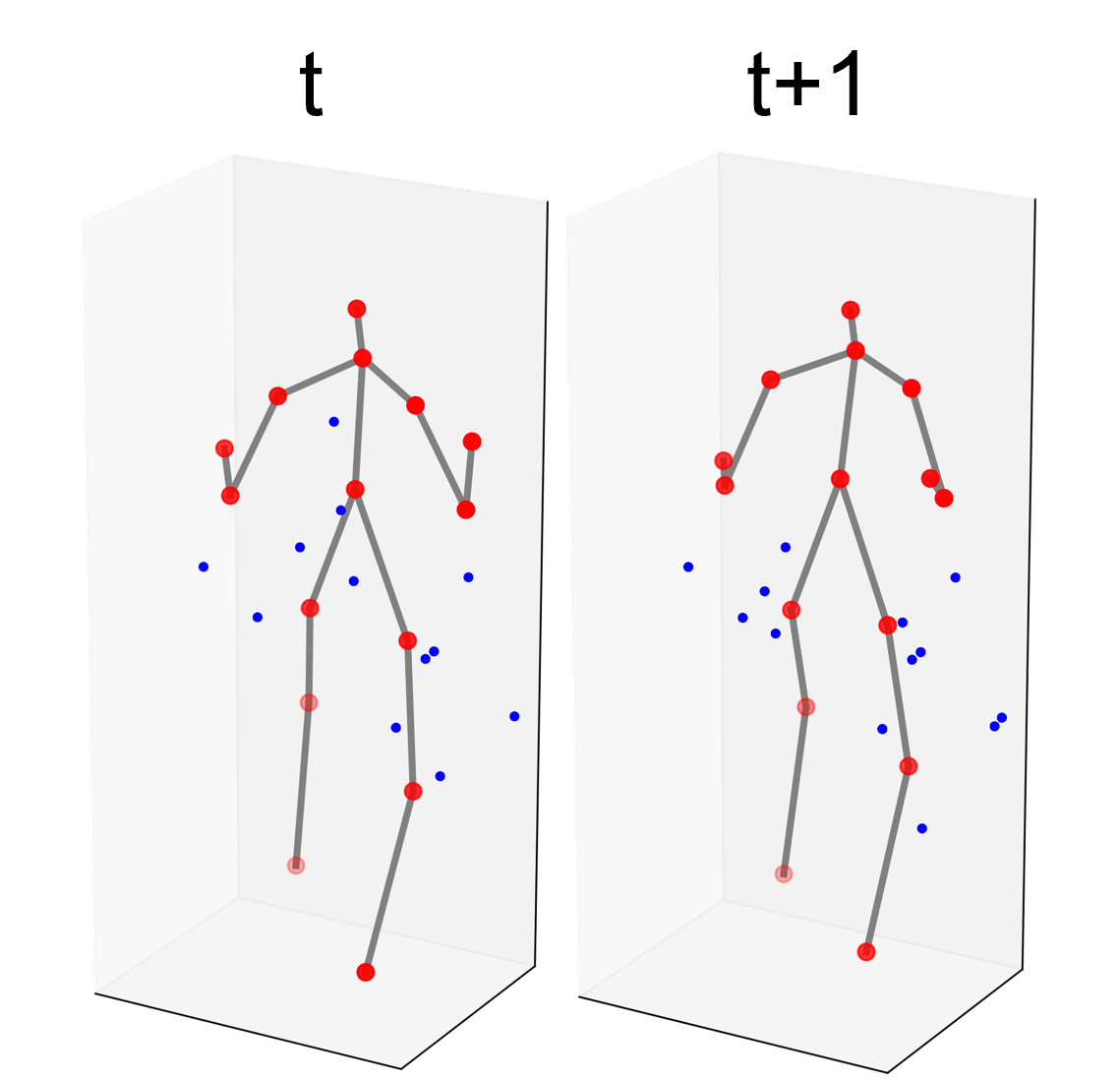}
        \caption{No UTCL}
    \end{subfigure}
    \begin{subfigure}[t]{.325\linewidth}
        \centering
        \includegraphics[width=\linewidth]{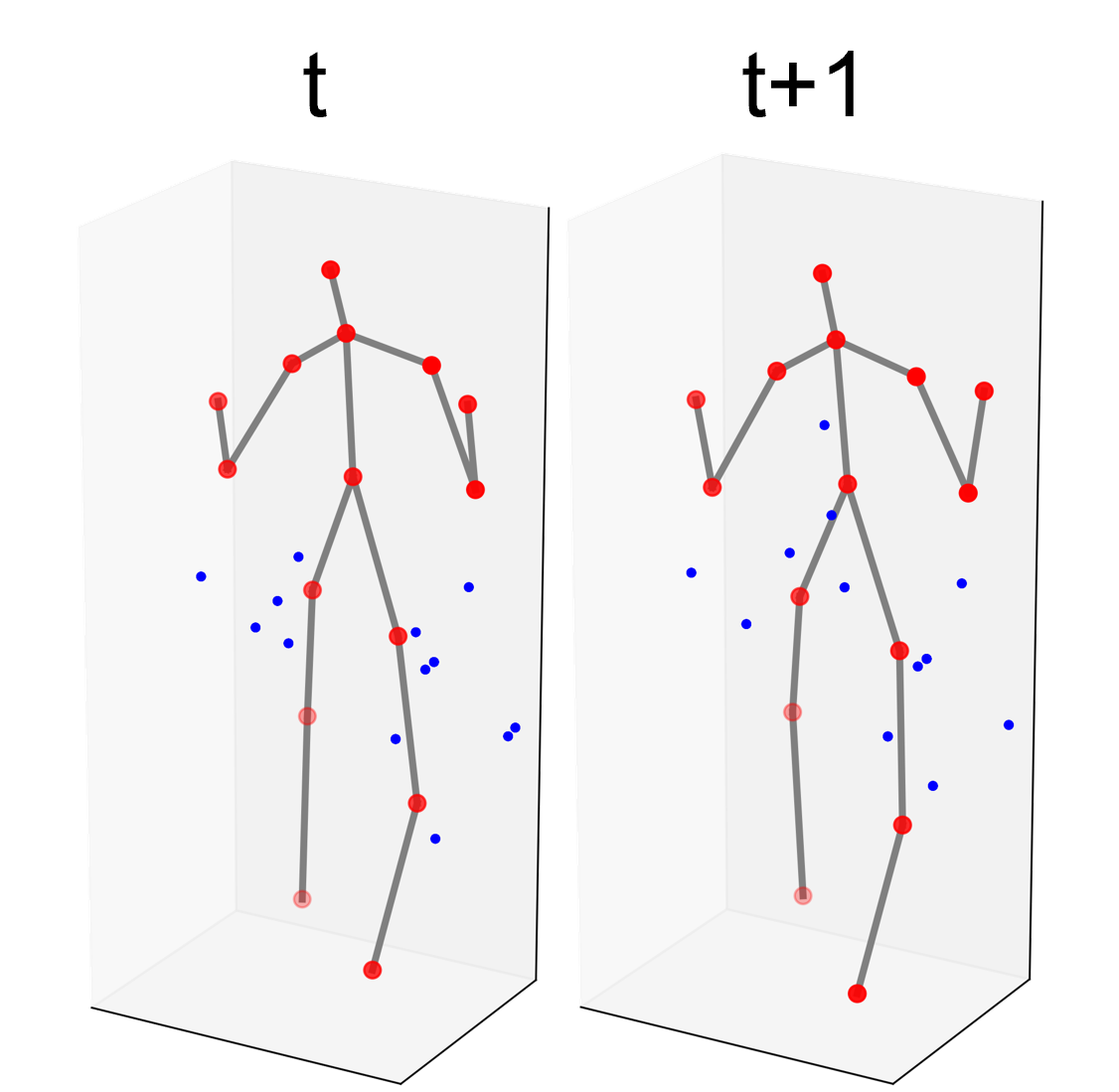}
        \caption{Ours}
    \end{subfigure}

    \caption{Comparison of pseudo-labels generated with and without UTCL. (a) Two consecutive ground-truth skeletons in $D^{\text{unlab}}$. (b) Pseudo-labels generated without using UTCL, (c) Pseudo-labels generated by \schname{} using UTCL. }
    \label{fig:pl}
\end{figure}

\begin{table}[t]
  \centering
  \small
  \caption{Ratio of PCs classified as mmWave in a binary classification task distinguishing between and mmWave and LiDAR data.}
  \begin{tabular}{c|cc|cc}
    \toprule
    Dataset & $D^{\text{mm}}$ & $D^{\text{lidar}}$ & $D^{\text{conv}}$ w/o FPF & $D^{\text{conv}}$ w/ FPF \\
    \midrule
    Ratio (\%) & 99.61 & 2.66 & 43.06 & \textbf{60.46}\\
  \bottomrule
  \end{tabular}
  \label{tab:cls}
\end{table}

\begin{table}[t]
  \centering
  \small
  \caption{Performance under different ratios of labeled mmWave data ($D^{\text{lab}}$ / MM-Fi). }
  \resizebox{\linewidth}{!}{
  \begin{tabular}{c|c|cc|cc}
    \toprule
    \multicolumn{2}{c|}{Setting} & \multicolumn{2}{c|}{F $\to$ F} & \multicolumn{2}{c}{F $\to$ B} \\
    \midrule
    Method & \makecell{Labeled\\Ratio} & MPJPE & \makecell{PA-\\MPJPE} & MPJPE & \makecell{PA-\\MPJPE} \\
    \midrule
    MT  & \multirow{2}{*}{1\%} & 18.40 & 13.20 & 19.94 & 12.23  \\
    \textbf{\schname} &        & \textbf{14.77} & \textbf{10.37} & \textbf{15.95} & \textbf{11.46}  \\
    \midrule
    MT  & \multirow{2}{*}{10\%}& 10.37 &  \textbf{6.80} & 16.97 & 12.12  \\
    \textbf{\schname} &        & \textbf{10.06} &  7.01 & \textbf{14.89} & \textbf{11.11}  \\
    \midrule
    MT  & \multirow{2}{*}{50\%}&  8.25 &  \textbf{5.63} & 17.02 & 12.15  \\
    \textbf{\schname} &        &  \textbf{8.15} &  5.68 & \textbf{15.61} & \textbf{11.46}  \\
    
  \bottomrule
  \end{tabular}}
  \label{tab:exp_ratio}
\end{table}


\begin{description}[style=unboxed, leftmargin=0cm]

\item[Preliminary study on point features:] 

We conduct a preliminary study to compare two point-level features, Doppler speed and height, under F $\to$ B. Each experiment is repeated with five different random seeds, and the results are averaged. As shown in~\cref{tab:feature}, using height achieves the lower error, indicating better cross-dataset generalization.

\item[Ablation on pseudo-labeling:] 

We investigate the impact of our pseudo-labeling strategy to utilize unlabeled mmWave data for dataset expansion. 
Results in~\cref{tab:pl} show that training the pseudo-label estimator on both $D^{\text{lab}}$ (with $L^{\text{lab}}$) and $D^{\text{unlab}}$ (with UTCL: DCL $L^{\text{dyn}}$ and SCL $L^{\text{sta}}$) significantly reduces error. 
Although the individual UTCL components, DCL and SCL, do not independently improve out-of-domain generalization, their combination yields substantial gains with only marginal in-domain error increase. This is because DCL or SCL alone biases predictions toward overly dynamic or static outcomes, while their combination maintains a better balance.

\item[Ablation on PC conversion:] 
We evaluate the effectiveness of our PC conversion pipeline in transforming a LiDAR dataset into its mmWave counterpart. As shown in ~\cref{tab:fpf}, each augmentation technique
incrementally improves performance. However, the most substantial gain is observed when our proposed FPF is included.

\item[Effects of hyperparameters:] 
We investigate the effects of key hyperparameters in UTCL ($\mu$, $\eta$, and $\rho$) and FPF ($\gamma$ and $\delta$), as shown in~\cref{tab:exp_hyper}. Each hyparameter is adjusted while the others are fixed. The model achieves optimal results with our selected hyperparameters under the F $\to$ B setting. 




\item[Visualization of pseudo-labels: ]
We visualize the pseudo-labels generated with or without our UTCL in~\cref{fig:pl}.
As illustrated, predictions without UTCL exhibit noticeable temporal inconsistencies, particularly in the hand regions. In contrast, our UTCL effectively enforces temporal consistency, resulting in smoother and more accurate pseudo-labels that closely align with the ground truth.

\item[Quantitative Analysis of converted PCs: ] 
To quantitatively assess how realistic the converted LiDAR PCs are, we train a binary classifier based on P4T~\cite{fanPoint4DTransformer2021} to distinguish between mmWave and LiDAR PCs. As reported in~\cref{tab:cls}, 60.46\% of converted LiDAR PCs are classified as mmWave data when FPF is employed, compared to only 43.06\% without FPF, demonstrating that our generated mmWave PCs are realistic and highlighting the importance of simulating the motion detection mechanism using FPF.


\item[Results with different labeled data availability:] 
\cref{tab:exp_ratio} examines the performance of \schname\ under varying ratios of $D^{\text{mm}}$ over the entire MM-Fi training set. The results indicate that \schname\ consistently outperforms the MT approach across all data ratios.
Notably, the most substantial error reduction is observed at 1\% labeled data, demonstrating the effectiveness of \schname\ in addressing data scarcity.



\end{description}

\section{Conclusion}
We propose \schname, a novel mmWave training approach to address data scarcity and limited diversity in mmWave HPE training by effectively utilizing unlabeled mmWave data and annotated LiDAR dataset. \schname\ expands an mmWave dataset through two key components: a pseudo-label estimator trained with an unsupervised temporal consistency loss to generate reliable pseudo-labels for unlabeled mmWave data, and a PC conversion method to convert LiDAR PCs into its mmWave counterparts by simulating mmWave PC attributes, including flow-based point filtering to simulate motion detection. 
Augmented with pseudo-labeled mmWave data and both converted LiDAR dataset, the original mmWave dataset is substantially expanded in volume and diversity. 
Experiments on multiple mmWave and LiDAR datasets show that models trained with \schname\ substantially outperform those trained on the original mmWave dataset alone, achieving superior accuracy and generalization across both in-domain and out-of-domain scenarios.




\section{Acknowledgement}
This work was supported, in part, by Research Grants Council
Collaborative Research Fund (under grant number C1045-23G)
and RGC-General Research Fund (under grant number 16201625) of Hong Kong.

{
    \small
    \bibliographystyle{ieeenat_fullname}
    \bibliography{main}
}

\clearpage
\setcounter{page}{1}
\maketitlesupplementary

\section{More Implementation Details}
\label{sec:imp}

This section presents more implementation details of our proposed \schname\ and a comparison scheme adapted for mmWave HPE. 

\begin{table}[t]
  \centering
  \small
  \caption{The entire PC conversion process in our \schname.}
  \resizebox{\linewidth}{!}{
  \begin{tabular}{c|c|l}
    \toprule
    Component & Parameters & Explanation \\
    \midrule
    \makecell{NPA} & \makecell{$\sigma_1=2\, \text{cm}$, \\$p=0.5$, \\$n=32$} & \makecell[l]{$n$ points sampled from\\ $\mathcal{N}(C, \sigma_1^2I)$ to a portion of $p$ \\of the LiDAR PCs, where $C$ is \\the center of the skeleton and $I$ \\is a $3\times 3$ identity matrix.} \\
    \midrule
    FPF & \makecell{$\gamma=2\,\text{cm}$, \\$\delta=10\,\text{cm}$} & \makecell[l]{The flow threshold is sampled \\from $\text{U}[\gamma, \delta]$.} \\
    \midrule
    \makecell{RS} & \makecell[l]{$r_{\text{min}}=0.125$\\$r_{\text{max}}=1.0$\\$m=128$} & \makecell[l]{A fraction of $r\in U[r_{\text{min}}, r_{\text{max}}]$\\ points are randomly sampled \\from the PC if it contains at \\least $m$ points. } \\
    \midrule
    \makecell{NI} & $\sigma_2=5\,\text{cm}$ & \makecell[l]{A noise  following $N(O, \sigma_2^2I)$ \\is injected to each point, where \\$O$ is the origin and $I$ is a $3\times 3$\\ identity matrix. }\\
    
  \bottomrule
  \end{tabular}
  }
  \label{tab:supple_syn}
\end{table}

\subsection{PC Conversion Pipeline}
We specify the parameters used in our PC conversion pipeline in \cref{tab:supple_syn}. The parameters are chosen based on empirical results on the validation set. $\upsilon$ is re-sampled per instance.

\subsection{Model Training}
The input PCs to the pseudo-label estimator and the inference HPE model are first normalized by subtracting the $(\tilde{X}, \tilde{Y}, \min{Z})$, where $\tilde{X}$, $\tilde{Y}$ are the medians of all X and Y coordinates of the skeleton sequence, and $\min{Z}$ is the minimum Z (height) coordinate among all points in the skeleton sequence. Outlier points are subsequently removed using a box filter with a range of $[-1.5\,\text{m}, 1.5\,\text{m}]$ along the X and Y axes, and $[0\,\text{m}, 2.0\,\text{m}]$ along the Z axis. After outlier removal, we apply the following augmentations during training: random rotation along the vertical axis within $[-10^\circ, 10^\circ]$, random scaling within $[0.9, 1.1]$, and random translation within $[-1\,\text{cm}, 1\,\text{cm}]$ along each axis. Finally, each PC is processed to 256 points through repetitive padding or truncation. In the learning rate scheduler, the duration of linear warmup is 20 epochs. 

\subsection{Adapted Mean-Teacher Pseudo-Labeling}
Mean-Teacher (MT) Pseudo-Labeling~\cite{tarvainenMeanTeachersAre2017} is a commonly used semi-supervised learning method designed for classification. We adapt it to mmWave HPE as a comparison scheme. The pseudo-label estimator (mean-teacher model) is an exponential moving average of the inference HPE model, with a decay rate of 0.999. The same augmentations are applied to the inputs of both the mean-teacher model and the inference model during pseudo-labeling.

\section{More Experimental Results}
In this section, we show more quantitative and visualization results for \schname.


\begin{table}[t]
  \centering
  \small
  \caption{Ablation studies on the LiDAR data size, represented by the ratio of data used in HmPEAR.}
  \resizebox{\linewidth}{!}{
  \begin{tabular}{c|c|cc|cc}
    \toprule
    \multicolumn{2}{c|}{Setting} & \multicolumn{2}{c|}{F $\to$ F} & \multicolumn{2}{c}{F $\to$ B} \\
    \midrule
    Method & \makecell{Labeled\\Ratio} & MPJPE & \makecell{PA-\\MPJPE} & MPJPE & \makecell{PA-\\MPJPE} \\
    \midrule
    MT  & \multirow{2}{*}{10\%}& 11.84 &  8.13 & 16.29 & 12.14  \\
    \textbf{\schname} &        & \textbf{10.59} &  \textbf{7.56} & \textbf{15.64} & \textbf{12.07}  \\
    \midrule
    MT  & \multirow{2}{*}{50\%}& 10.65 &  7.40 & 16.26 & \textbf{12.05}  \\
    \textbf{\schname} &        & \textbf{10.27} &  \textbf{7.19} & \textbf{16.25} & 12.55 \\
    \midrule
    MT  &\multirow{2}{*}{100\%}& 10.37 &  \textbf{6.80} & 16.97 & 12.12  \\
    \textbf{\schname} &        & \textbf{10.06} &  7.06 & \textbf{14.89} & \textbf{11.11}  \\
    
  \bottomrule
  \end{tabular}}
  \label{tab:exp_lidarratio}
\end{table}

\begin{table}[t]
    \centering
    \small
    \caption{Ablation study on PC conversion of a LiDAR dataset.
    } 
    \begin{tabular}{cccc|cc}
      \toprule
      \multicolumn{4}{c|}{Components} & \multicolumn{2}{c}{F $\to$ B} \\
      \midrule
      NPA & FPF & RS & NI & MPJPE & PA-MPJPE \\
      \midrule
                &           &           &           & 15.85 & 12.23 \\
      \midrule
      \ding{51} &           &           &           & 15.80 & 12.23 \\
                & \ding{51} &           &           & 15.63 & 11.80 \\
                &           & \ding{51} &           & 15.63 & 12.01 \\
                &           &           & \ding{51} & 15.84 & 12.22 \\
      \midrule
      \ding{51} & \ding{51} & \ding{51} &           & 15.47 & 11.56 \\
      \ding{51} & \ding{51} &           & \ding{51} & 15.70 & 11.69 \\
      \ding{51} &           & \ding{51} & \ding{51} & 15.54 & 11.75 \\
                & \ding{51} & \ding{51} & \ding{51} & 15.37 & 11.58 \\
      \midrule
      \ding{51} & \ding{51} & \ding{51} & \ding{51} & \textbf{14.89} & \textbf{11.11} \\
      \bottomrule
    \end{tabular}
    \label{tab:fpf_complete}
\end{table}


\subsection{Complete Ablation Study on PC Conversion}
We present a more complete ablation study on each component of our PC conversion pipeline in~\cref{tab:fpf_complete}. Results show that solely using each individual component brings performance gains, while removing any component from the full pipeline degrades performance. This demonstrates that each component plays a vital role in effectively converting LiDAR PCs to simulate mmWave PCs.

\subsection{Results with different LiDAR data availability} 
To evaluate \schname\ under varying LiDAR data availability, we conduct experiments using a different ratio of data from HmPEAR. Results in~\cref{tab:exp_lidarratio} show that \schname\ significantly improves performance even with only 10\% of HmPEAR, demonstrating its efficiency in leveraging LiDAR data.

\begin{figure}[t]
    \centering
    \includegraphics[width=\linewidth]{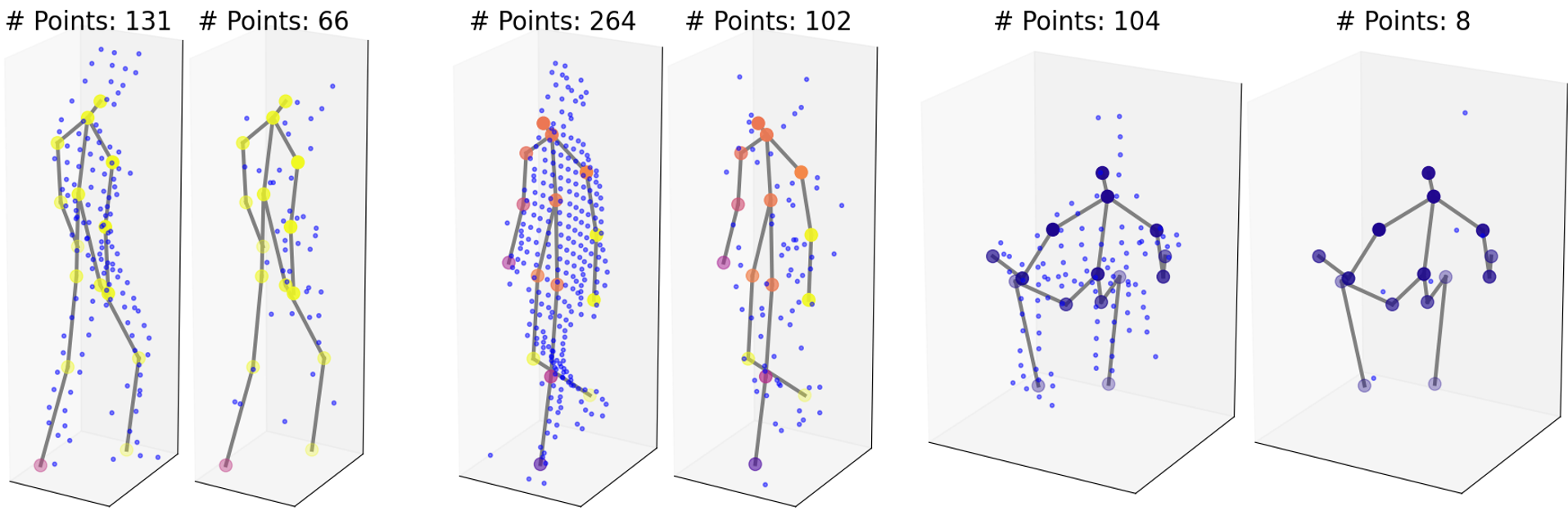}

    \caption{More visualization results of point cloud conversion. Left: original LiDAR PCs. Right: converted mmWave PCs. Joints with high flow values are yellow, while those with low flow values are blue.}
    \label{fig:pcc}
\end{figure}

\begin{figure}[t]
    \centering
    \begin{subfigure}[t]{.24\linewidth}
        \centering
        \includegraphics[width=\linewidth]{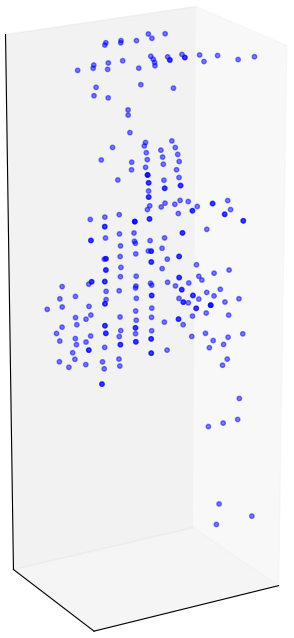}
    \end{subfigure}
    \begin{subfigure}[t]{.24\linewidth}
        \centering
        \includegraphics[width=\linewidth]{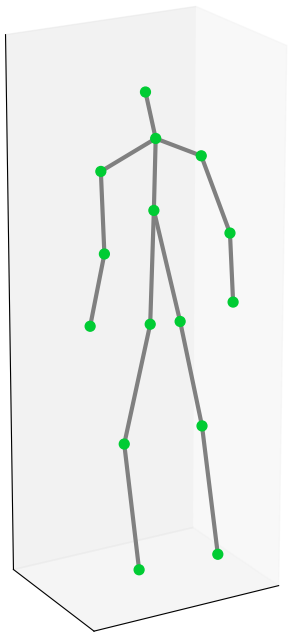}
    \end{subfigure}
    \begin{subfigure}[t]{.24\linewidth}
        \centering
        \includegraphics[width=\linewidth]{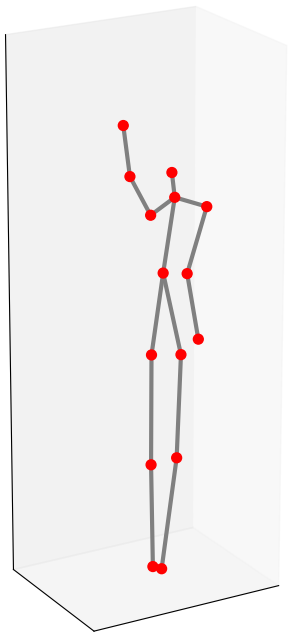}
    \end{subfigure}
    \begin{subfigure}[t]{.24\linewidth}
        \centering
        \includegraphics[width=\linewidth]{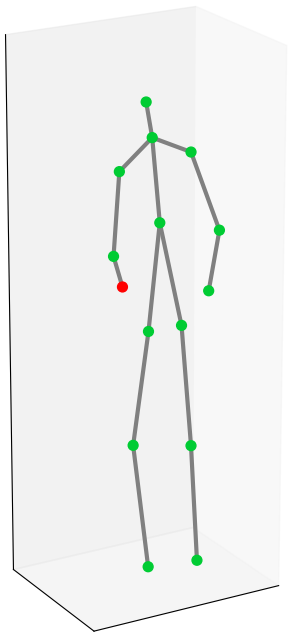}
    \end{subfigure}
    \begin{subfigure}[t]{.24\linewidth}
        \centering
        \includegraphics[width=\linewidth]{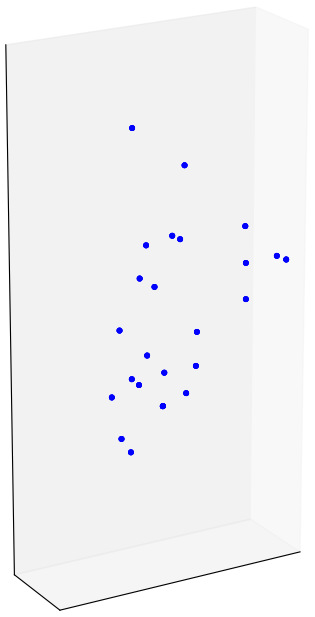}
    \end{subfigure}
    \begin{subfigure}[t]{.24\linewidth}
        \centering
        \includegraphics[width=\linewidth]{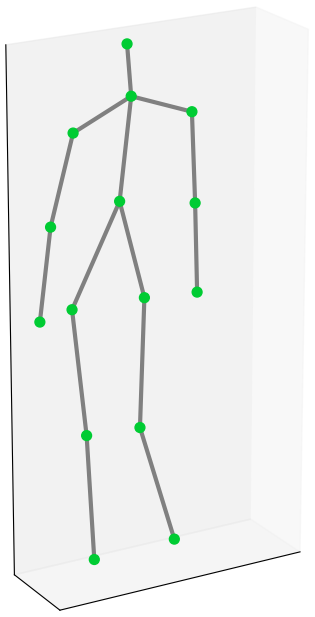}
    \end{subfigure}
    \begin{subfigure}[t]{.24\linewidth}
        \centering
        \includegraphics[width=\linewidth]{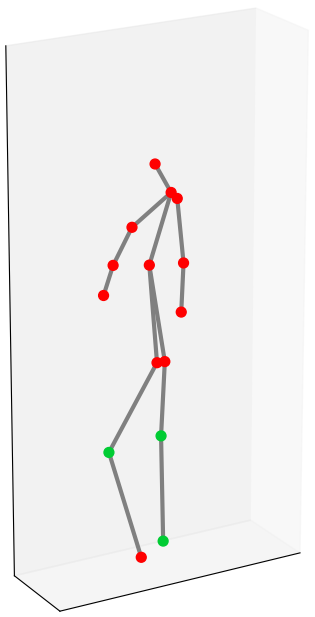}
    \end{subfigure}
    \begin{subfigure}[t]{.24\linewidth}
        \centering
        \includegraphics[width=\linewidth]{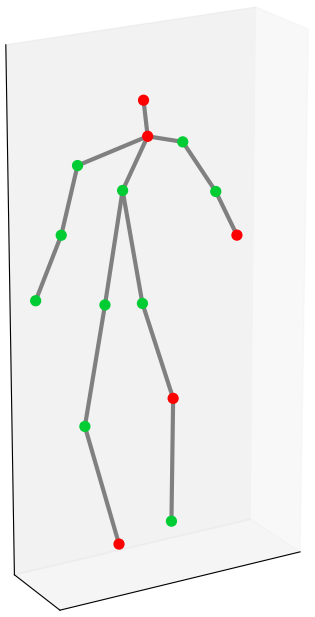}
    \end{subfigure}
    \begin{subfigure}[t]{.24\linewidth}
        \centering
        \includegraphics[width=\linewidth]{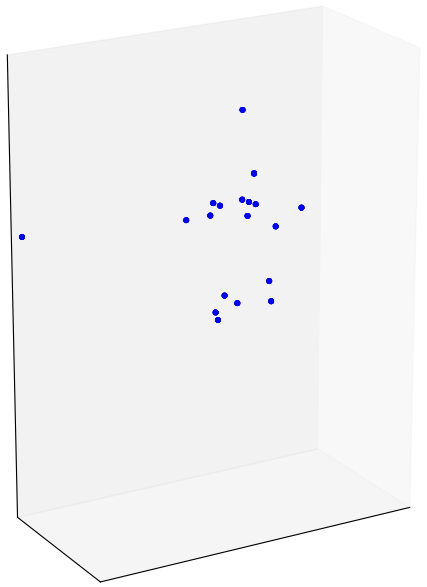}
        \caption{Input PC}
    \end{subfigure}
    \begin{subfigure}[t]{.24\linewidth}
        \centering
        \includegraphics[width=\linewidth]{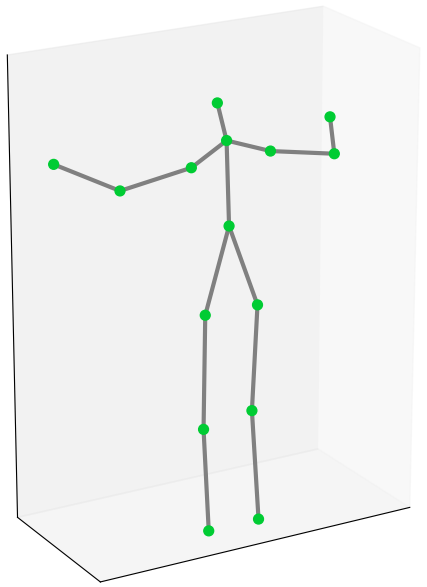}
        \caption{GT}
    \end{subfigure}
    \begin{subfigure}[t]{.24\linewidth}
        \centering
        \includegraphics[width=\linewidth]{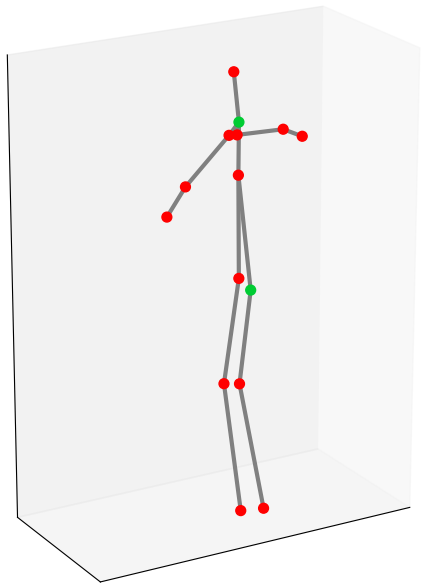}
        \caption{P4T}
    \end{subfigure}
    \begin{subfigure}[t]{.24\linewidth}
        \centering
        \includegraphics[width=\linewidth]{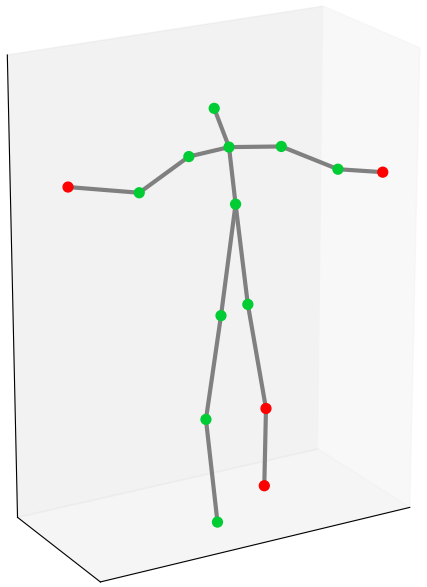}
        \caption{Ours}
    \end{subfigure}

    \caption{Visualization results on MM-Fi $\rightarrow$ mmBody (first row) and mmBody $\rightarrow$ MM-Fi (second and third rows). Joints with high error ($>10\,\text{cm}$) are colored red while others are colored green. }
    \label{fig:vis}
\end{figure}

\subsection{More Visualization on PC Conversion}
We provide more visualization results of our PC conversion pipeline in \cref{fig:pcc}. It can be observed that the converted PCs exhibit increased noisiness, reduced point density, with more points gathering around fast moving joints, effectively simulating mmWave PC attributes.

\subsection{More Visualizations on HPE Results}
\cref{fig:vis} shows more visualization results for \schname. The first row compares P4T predictions trained on MM-Fi without versus with \schname\ (augmented by HmPEAR), while subsequent rows compare models trained on mmBody~\cite{chenMmBodyBenchmark3D2023} expanded with LiDARHuman26M~\cite{liLiDARCapLongrangeMarkerless2022}. It is clearly shown that using \schname\ leads to consistently higher performance, even on sparse and noisy PCs. 

\section{Limitation and Future Work}
While \schname\ significantly improves performance by expanding mmWave datasets with unlabeled data and LiDAR datasets, it has certain limitations that pave the way for future research. 
First, the PC conversion pipeline relies on empirical parameter settings, which may not be optimal for all scenarios. Future work could explore adaptive or learnable conversion methods for better simulating mmWave PC attributes from different LiDAR datasets.
Second, while UCTL effectively encourages temporal consistency, it may not fully capture complex motion patterns. Future research could investigate more sophisticated temporal modeling techniques, further refining the pseudo-label quality.
Lastly, \schname\ currently focuses on single-person HPE; extending it to multi-person scenarios would be a valuable direction, which includes addressing challenges such as occlusion and interaction between multiple subjects.

\end{document}